\documentclass[conference]{IEEEtran}
\usepackage{cite}
\usepackage{amsmath,amssymb,amsfonts}
\usepackage{algorithmic}
\usepackage{graphicx}
\usepackage{textcomp}
\usepackage{xcolor}
\def\BibTeX{{\rm B\kern-.05em{\sc i\kern-.025em b}\kern-.08em
    T\kern-.1667em\lower.7ex\hbox{E}\kern-.125emX}}
    
\usepackage{multirow}
\usepackage{makecell}
\usepackage{mathtools}
\usepackage{mathrsfs}
\usepackage[caption=false,font=footnotesize]{subfig}		
\usepackage{tabu} 
\usepackage{xcolor} 
\usepackage{varwidth}
\usepackage[hyphens]{url}
\newsavebox\tmpbox

\graphicspath{ {images/} }
\newcommand{\bvec}[1]{\vec{\boldsymbol{#1}}}

\newcommand{\TODO}[1]{\iftrue \textcolor{blue}{TODO: #1} \fi}
\newcommand{\REVISE}[1]{\iftrue \textcolor{red}{#1} \fi}

\makeatletter

\def\ps@IEEEtitlepagestyle{
  \def\@oddfoot{\mycopyrightnotice}
  \def\@evenfoot{}
}
\def\mycopyrightnotice{
  {\footnotesize 978-1-7281-0858-2/19/\$31.00~\copyright~2019 IEEE\hfill} 
  \gdef\mycopyrightnotice{}
}
    
\begin{document}

\title{Demystifying Learning Rate Policies for \\ High Accuracy Training of Deep Neural Networks
}

\author{\vspace{0.8mm}\IEEEauthorblockA{\IEEEauthorblockN{Yanzhao Wu$^+$, Ling Liu$^+$, Juhyun Bae$^+$, Ka-Ho Chow$^+$,\\ Arun Iyengar$^*$, Calton Pu$^+$, Wenqi Wei$^+$, Lei Yu$^*$, Qi Zhang$^*$}
\IEEEauthorblockA{$^+$\textit{School of Computer Science, Georgia Institute of Technology, Atlanta, GA, USA}\\
$^*$\textit{IBM Thomas. J. Watson Research, Yorktown Heights, NY, USA}
}
}
\vspace{1.2mm}
}


\maketitle
\begin{abstract}
Learning Rate (LR) is an important hyper-parameter to tune for effective training of deep neural networks (DNNs). Even for the baseline of a constant learning rate, it is non-trivial to choose a good constant value for training a DNN. Dynamic learning rates involve multi-step tuning of LR values at various stages of the training process and offer high accuracy and fast convergence. However, they are much harder to tune. In this paper, we present a comprehensive study of 13 learning rate functions and their associated LR policies by examining their range parameters, step parameters, and value update parameters. We propose a set of metrics for evaluating and selecting LR policies, including the classification confidence, variance, cost, and robustness, and implement them in LRBench, an LR benchmarking system. LRBench can assist end-users and DNN developers to select good LR policies and avoid bad LR policies for training their DNNs. We tested LRBench on Caffe, an open source deep learning framework, to showcase the tuning optimization of LR policies. Evaluated through extensive experiments, we attempt to demystify the tuning of LR policies by identifying good LR policies with effective LR value ranges and step sizes for LR update schedules.
\end{abstract}


\section{Introduction}
Deep neural networks (DNNs) are widely employed to mine Big Data and gain deep insight on Big Data, ranging from image classification, voice recognition, text mining and Natural Language Processing (NLP). One of the most important performance optimization for DNNs is to train a deep learning model capable of achieving high test accuracy.

Training a deep neural network (DNN) is an iterative global optimization problem. During each iteration, a loss function is employed  
to measure the deviation of its prediction to the ground truth and update the model parameters $\theta$, aiming to minimize this loss function $L_\theta$. Gradient Descent (GD) is a popular class of non-linear optimization algorithms that iteratively learn to minimize such a loss function~\cite{mnistlenet,alexnet,clr-22,clr-20,resnet}. Example GD optimizers include Stochastic GD~\cite{sgd}, Momentum~\cite{momentum}, Nesterov~\cite{nesterov}, Adagrad~\cite{adagrad}, AdaDelta~\cite{adadelta}, Adam~\cite{adam} and so forth. We consider SGD for updating the parameters $\theta$ of a DNN by Formula (\ref{formula:sgd}): 
\begin{equation}
\theta_{t} = \theta_{t-1} - \eta_{t} \nabla \boldsymbol{L}_\theta
\label{formula:sgd}
\end{equation}
where $\boldsymbol{L}_\theta$ is the loss function, $\nabla \boldsymbol{L}_\theta$ represents the gradients, $t$ is the current iteration, and $\eta_{t}$ is the learning rate, which controls the extent of the update to the parameters $\theta$ at iteration $t$. Choosing a proper learning rate can be difficult. A too small learning rate may lead to slow convergence, while a too large learning rate can deter convergence and cause the loss function to fluctuate and get stuck in a local minimum or even to diverge~\cite{Bengio-PracticalRecommendations,clr-1,clr-3}. 
Due to the difficulty of determining a good LR policy, the constant LR is a baseline default LR policy for training DNNs in several deep learning (DL) frameworks (e.g., Caffe, TensorFlow, Torch), numerous efforts have been engaged to enhance the constant LR by incorporating a multi-step dynamic learning rate schedule, which attempts to adjust the learning rate during different stages of DNN training by using a certain type of annealing techniques. However, good LR schedules need to adapt to the characteristics of different datasets and/or different neural network models~\cite{dawnbench,GTDLBenchICDCS,GTDLBenchBigData,GTDLBenchTSC}. 
Empirical approaches are used manually in practice to find good LR values and update schedules through trials and errors. Moreover, due to the lack of relevant systematic studies and analyses, the large search space for LR parameters often results in huge costs for this hand-tuning process, impairing the efficiency and performance of DNN training.

\begin{table*}[h!]
\caption{\small{Definition of the 13 Learning Rate Functions}}
\label{table:lr}
\centering
\scalebox{1.0}{
\footnotesize
\begin{tabu}{|c|c|c|c|c|c|c|c|}
\hline
\multicolumn{2}{|c|}{Learning Rates} & abbr. & $k_0$ & $k_1$ & $g(t)$ & Schedule & \#Param  \\
\hline
\multicolumn{2}{|c|}{Fixed} & FIX & $k_0 $ & $k_0$ & N/A & fixed & 1 \\
\hline
\multirow{5}{*}{\makecell{Decaying LR}} & \makecell{fixed step\\size}  & STEP & $k_0$ & 0 & $\gamma ^{floor(t/l)}$ & $t,l$ & 3 \\
\cline{2-8}
& \makecell{variable step\\sizes} & NSTEP & $k_0$ & 0 & \makecell{Given $n$ step sizes: $l_0, l_1, l_2, ..., l_{n-1}$, \\ $\gamma ^i, \; i\in \mathbb{N}~s.t.~l_{i-1} \le t < l_i$} & $t,l_i$ &  $n+2$ \\
\cline{2-8}
 & \makecell{exponential\\function} & EXP & $k_0$ & 0 & $\gamma ^ t$ & $t$ & 2 \\
\cline{2-8}
& \makecell{inverse\\time} & INV & $k_0$ & 0 & $\frac{1}{(1+t\gamma)^p}$ & $t$ & 3 \\
\cline{2-8}
& \makecell{polynomial\\function} & POLY & $k_0$ & 0  & $(1-\frac{t}{l})^p$ & \makecell{$t,l$ \\ ($l = max\_iter$)} & 4 (3) \\
\hline
\multirow{7}{*}{CLR} & triangular & TRI & $k_0$ & $k_1$ & \makecell{$TRI(t)=\frac{2}{\pi}|arcsin(sin(\frac{\pi t}{2l}))|$} & $t, l$ & 3 \\
\cline{2-8}
 & triangular2 & TRI2 & $k_0$ & $k_1$ & $\frac{1}{2^{floor(\frac{t}{2l})}}TRI(t)$ & $t, l$ & 3 \\
\cline{2-8}
 & triangular\_exp & TRIEXP & $k_0$ & $k_1$ & $\gamma^{t}TRI(t)$ & $t, l$ & 4 \\
\cline{2-8}
& sin & SIN & $k_0$ & $k_1$ & \makecell{$SIN(t)=|sin(\pi\frac{t}{2l})|$} & $t, l$ & 3 \\
\cline{2-8}
 & sin2 & SIN2 & $k_0$ & $k_1$ & \makecell{$\frac{1}{2^{floor(\frac{t}{2l})}}SIN(t)$} & $t, l$ & 3 \\
\cline{2-8}
 & sin\_exp & SINEXP & $k_0$ & $k_1$ & \makecell{$\gamma^{t}SIN(t)$} & $t, l$ & 4 \\
\cline{2-8}
& cosine & COS & $k_0$ & $k_1$ & \makecell{$COS(t)=\frac{1}{2}(1+cos(\pi \frac{2t}{l}))$} & $t, l$ & 3 \\
\hline
\end{tabu}
}
\vspace{-4mm}
\end{table*}
In this paper, we present a comprehensive study of 13 learning rate functions and their associated LR policies by examining their range parameters, step parameters, and value update parameters. We propose a set of metrics for evaluating, comparing and selecting good LR policies in terms of LR utility, robustness and cost, and avoiding bad LR policies. We develop a LR benchmarking system, called LRBench, to support automated or semi-automated evaluation and selection of good LR policies for DNN developers and end-users. To demystify LR tuning, we implement all 13 LR functions and the LR policy evaluation metrics in LRBench, and provide analytical and experimental guidelines for identifying and selecting good policies for each LR function and illustrate why some LR policies are not recommended. We conduct comprehensive experiments using MNIST and CIFAR-10 and three popular neural networks: LeNet~\cite{mnistlenet}, a 3-layer CNN (CNN3)~\cite{caffe} and ResNet~\cite{resnet-32}, to demonstrate the accuracy improvement by choosing good LR policies, avoiding bad ones, and the impact of datasets and DNN model specific characteristics on LR tuning.

\section{Characterization of Learning Rates}
Learning rate ($\eta$) as a global hyperparameter determines the size of the steps which a GD optimizer takes along the direction of the slope of the surface derived from the loss function downhill until reaching a (local) minimum (valley). Small LRs will slow down the training progress, but are also more desirable when the loss keeps getting worse or oscillates wildly. Large LRs will accelerate the training progress rate and are beneficial when the loss is dropping fairly consistently in several consecutive iterations. However, large LR value may also cause the training getting stuck at a local minimum, preventing the model from improving, resulting in either a low accuracy or not able to converge. Ideally, one wishes to have a suitable LR policy that will gradually improve the model and lead to a high accuracy at the end of the training. However, finding such an optimal LR is difficult. Typically, numerous experiments need to be performed with trials and errors to find the best values. Thus, manually tuning to find a good LR policy is rare. Using a very small constant learning rate (e.g., 0.01 for MNIST and 0.001 for CIFAR-10) becomes a {\em not-bad} default LR policy.

\begin{figure*}[h!]
\centering
\subfloat[At the 70th iteration]{
  \centering
  \includegraphics[trim=60 25 0 25, clip,width=0.33\textwidth]{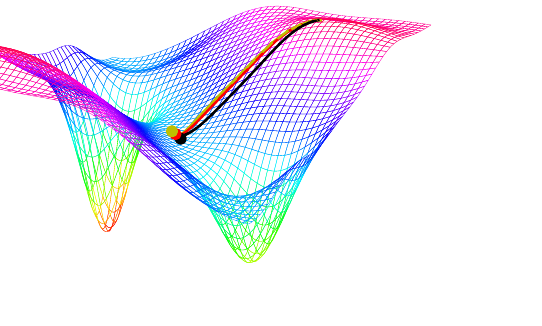}
  \label{fig:fix-nstep-triexp-70}
} 
\hspace{-6mm}
\subfloat[At the 115th iteration]{
  \centering
  \includegraphics[trim=60 25 0 25, clip,width=0.33\textwidth]{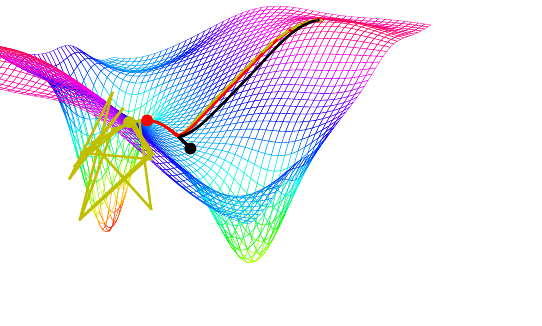}
  \label{fig:fix-nstep-triexp-115}
} 
\hspace{-6mm}
\subfloat[At the 199th iteration]{
  \centering
  \includegraphics[trim=60 25 0 25, clip,width=0.33\textwidth]{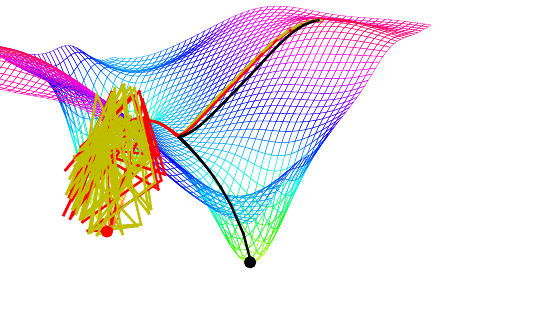}
  \label{fig:fix-nstep-triexp-199}
} 
\caption{\small{Visualization of the Training Process with Different LRs}}
\label{fig:visulization-fix-nstep-triexp}
\vspace{-6mm}
\end{figure*}

Several orthogonal and yet complimentary efforts have been engaged to address the problems of finding good LR values. Table \ref{table:lr} provides the definitions of 13 LR functions used in our study. Each LR function consists of one or more parameters. When the fixed LR sets its value to 0.01, we call $LR_{FIX}(k_0=0.01)$ an LR policy. Formula (\ref{formula:lr}) provides a general formulation:
\vspace{-1mm}
\begin{equation}
    \eta(t) = |k_0 - k_1| g(t) + min(k_0, k_1)
\label{formula:lr}
\vspace{-1mm}
\end{equation}
\noindent
where $t$ represents the current iteration, $k_0$, $k_1$ represent the learning rate refinement value range for a given LR step. LR typically starts from $k_0$, and its value is determined by $g(t) \in [0,1]$, an LR function that defines the specific rule for updating the LR values. We classify the 13 LR functions into three categories: fixed, decaying and cyclic.
(1)The {\bf fixed LR} is defined by $\eta = k_0$ and $k_1=k_0$, and it has only one parameter $k_0$ with a constant value from the domain of $(0,1)$. Although using a relatively small fixed LR value is a conservative approach of avoiding bad LR policies, it has
the high cost of slow convergence and long training time, combined with little guarantee of high accuracy, numerous efforts have been engaged in developing decaying LRs and cyclic LRs.
(2) {\bf Decaying Learning Rates\/} use different forms of decay functions to improve the convergence speed of DNN training by LR annealing. The STEP function defines the LR annealing schedule by a fixed step size, such as an update every $l$ iterations ($l>1$). It starts with a relatively large LR value, defined by $k_0$, and decays the LR value over time, for example, by an exponential factor $\gamma$ defined by the STEP function $g(t)$. Instead of the fixed step size, the NSTEP function uses variable step sizes, say $l_0, l_1, l_2, \dots, l_{n-1}$. The LR value is initialized by $k_0$ (when $i=0$) and computed by $\gamma^i$ (when $i>0$ and $l_{i-1}\leq t\le l_i$) (recall Table \ref{table:lr}). Note that the variable step sizes are pre-defined annealing schedules for decaying the LR value over time.
For other decaying LRs, $g(t)$ is a decaying function, such as the exponential function, the inverse time function, and the polynomial function. All decaying LRs can be formulated simply as $\eta(t) = k_0 g(t)$, by setting $k_1=0$.
(3) {\bf Cyclic Learning Rates (CLRs)\/} vary the LR value within a pre-defined value range cyclically during each predefined LR step, instead of setting it to a fixed value or using some form of decay schedule.~\cite{superconvergence} shows that CLR can train a DNN faster to reach a certain accuracy threshold. Three types of CLR methods are considered in our study: triangle-based, sin-based and cosine-based.~\cite{clr} proposed triangle-based CLRs with periodic schedule, cut-in-half cyclic decay schedule (TRI2) and exponential decay schedule (TRIEXP).~\cite{sgdr} proposed a variant of triangle based CLRs using the cosine function. We also implement three sin-based CLR functions: sin, sin2 and sin\_exp, corresponding to the TRI, TRI2 and TRIEXP.
For each cyclic LR function, one needs to configure three important parameters in advance to produce a good LR policy. The first one is the cycle length ($l$), which is defined by the half of the cycle period to correspond to the step size $l$ in decaying LRs. 
The second and third parameters are the LR boundary values ($k_0, k_1$) that bound the LR value update by the update schedule during each cycle.


Figure~\ref{fig:visulization-fix-nstep-triexp} visualizes the optimization process of three LR functions: FIX, NSTEP and TRIEXP. 
The total number of iterations is 200 and all three LRs start from the same initial point. The corresponding LR policies are FIX($k_0=0.025$) in black, NSTEP($k_0=0.05,\gamma=0.1,l=[150,180]$) in red and TRIEXP($k_0=0.05,k_1=0.3,\gamma=0.94,l=100$) in yellow. 
We observe that at the beginning of optimization, TRIEXP achieved the fastest optimization progress as Figure \ref{fig:fix-nstep-triexp-70} and \ref{fig:fix-nstep-triexp-115} show. It shows that starting from a high LR value and decrease it as training progresses may help accelerate the training process.
Also different LR policies will result in different optimization paths. Even though the three LRs exhibit little difference w.r.t. their optimization paths up to the 70th iteration, FIX reaches a very different local optimum at the 199th iteration instead of the global optimum (red). This observation shows that the accumulated impact of the LR updates could also lead to sub-optimal results. 
It might be that high LRs introduce high ``kinetic energy'' into the optimization and thus model parameters are bouncing around chaotically as shown by the TRIEXP update path (yellow) in Figure \ref{fig:fix-nstep-triexp-115} and \ref{fig:fix-nstep-triexp-199}, where the model may not converge to an optimum.
 



\section{Evaluation Metrics for LR Policies}
In order to compare different LR values and schedules for each LR function, we propose three sets of metrics to evaluate and compare different LR policies in terms of utility, cost and robustness when the LR policy is used for training DNN.
We use the prediction accuracy and confidence of the trained model to measure the {\bf utility} of a specific LR policy with respect to the effectiveness of DNN training. We measure 
accuracy using Top-1 accuracy and Top-5 accuracy. 
{\bf Top-1 accuracy} considers the prediction a match only if the class with the highest prediction probability, i.e., $\boldsymbol{F}_\theta(X_i)$,  is the same as the ground truth label. {\bf Top-5 accuracy} relaxes this requirement such that the prediction is considered a match to the ground truth label as long as the ground truth label is among the predicted classes of the top 5 highest probabilities. Both are computed as the number of matched predictions divided by the number of samples evaluated.
We define the classification confidence in terms of the probability of the correctly classified sample matching the ground truth label ($y_i$), denoted by $\boldsymbol{P}_\theta(X_i)_{y_i}$, the $y_i^{th}$ element of the predicted probabilities (i.e., the softmax layer output). We measure the classification confidence and its deviation in terms of average confidence, confidence derivation and confidence derivation across classes.
%
%
%
%


{\bf Average Confidence (AC)} measures the average classification confidence of the correctly classified samples. It is defined as Formula (\ref{formula:metric-ac}) where $count()$ counts the number of samples satisfying the conditions within the parentheses.
\vspace{-2mm}
\begin{equation}
    \begin{aligned}
    AC(\boldsymbol{P}_\theta, \bvec{X}) &= \frac{1}{NS}(\sum_{i=1}^{n}\boldsymbol{P}_\theta(X_i)_{y_i} count(\boldsymbol{F}_\theta(X_i) = y_i)) \\
    NS &= \sum_{i=1}^{n}count(\boldsymbol{F}_\theta(X_i) = y_i)
    \end{aligned}
\label{formula:metric-ac}
\vspace{-1mm}
\end{equation}

{\bf Confidence Deviation (CD)} (Formula (\ref{formula:metric-cd})) is the standard deviation ($std()$) of the classification confidence of the correctly classified samples. Small deviations imply stable classification performance.
\vspace{-1mm}
\begin{equation}
CD(\boldsymbol{P}_\theta, \bvec{X}) = std_i(\boldsymbol{P}_\theta(X_i)_{y_i}) \quad s.t. \quad \boldsymbol{F}_\theta(X_i) = y_i
\label{formula:metric-cd}
\vspace{-1mm}
\end{equation}

\textbf{Confidence Deviation Across Classes (CDAC)} measures the degree of imbalance across prediction classes, since multinomial classification tends to suffer from imbalanced datasets, e.g., minority classes. If the trained model is tuned for majority classes, then it may not give fair prediction results for the minority classes~\cite{imbalanced-data}. The CDAC metric assesses the fairness of the trained DNN by measuring the standard deviation among the average confidence for each class $i$, denoted by $AC_i$, and it is defined as Formula (\ref{formula:metric-cdac}):
\vspace{-2mm}
\begin{equation}
    \begin{aligned}
    CDAC(\boldsymbol{P}_\theta, \bvec{X}) &= std_i(AC_i)\\     
    AC_i &= \frac{\sum_{j=1}^{n}\boldsymbol{P}_\theta(X_j)_i count(\boldsymbol{F}_\theta(X_j) = y_j = i)}{\sum_{j=1}^{n} count(\boldsymbol{F}_\theta(X_j) = y_j = i)}
    \end{aligned}
\label{formula:metric-cdac}
\vspace{-1mm}
\end{equation}

{\bf Cost and Robustness.\/} Two metrics measure the cost of tuning LRs, and one metric measures the robustness of LRs.
(1){\bf \#Parameters (\#Param)}: It is a metric that indirectly reflects the cost of tuning LR parameters to find a good LR policy. Recall Table \ref{table:lr}, the list of the \#Param for each LR function is about 3$\sim$4 except NSTEP, which is $n+2$. Table \ref{table:metric-evaluation-mnist}$\sim$\ref{table:metric-evaluation-cifar10-resnet} show that the good NSTEP policies have only 2 to 5 steps. 
(2) {\bf \#Iterations at the Highest Top-1 Accuracy}: It indicates at which iteration the highest accuracy is achieved. Our experiments show that the highest accuracy can be achieved often before the training reaches the pre-defined (default) total \#iterations. For a total of $L$ training iterations, we calculate the accuracy every $R$ ($R<<L$) iterations and use the highest accuracy as the measurement result. 
(3) {\bf Loss Difference (LD)}: It is a good indicator of the robustness 
against over-fitting. Over-fitting means that the trained model over-fits to the training dataset, and may result in low loss on the training dataset ($train\_loss$) but high loss on the testing dataset ($test\_loss$), leading to low accuracy. The Loss Difference (LD) between the $test\_loss$ and $train\_loss$ measures the impact of a specific LR policy to combat over-fitting, which is defined as Formula (\ref{formula:metric-ld}).
\begin{equation}
LD = test\_loss - train\_loss
\label{formula:metric-ld}
\end{equation}

\begin{figure}[h!]
\vspace{-6mm}
\centering
    \includegraphics[width=0.48\textwidth]{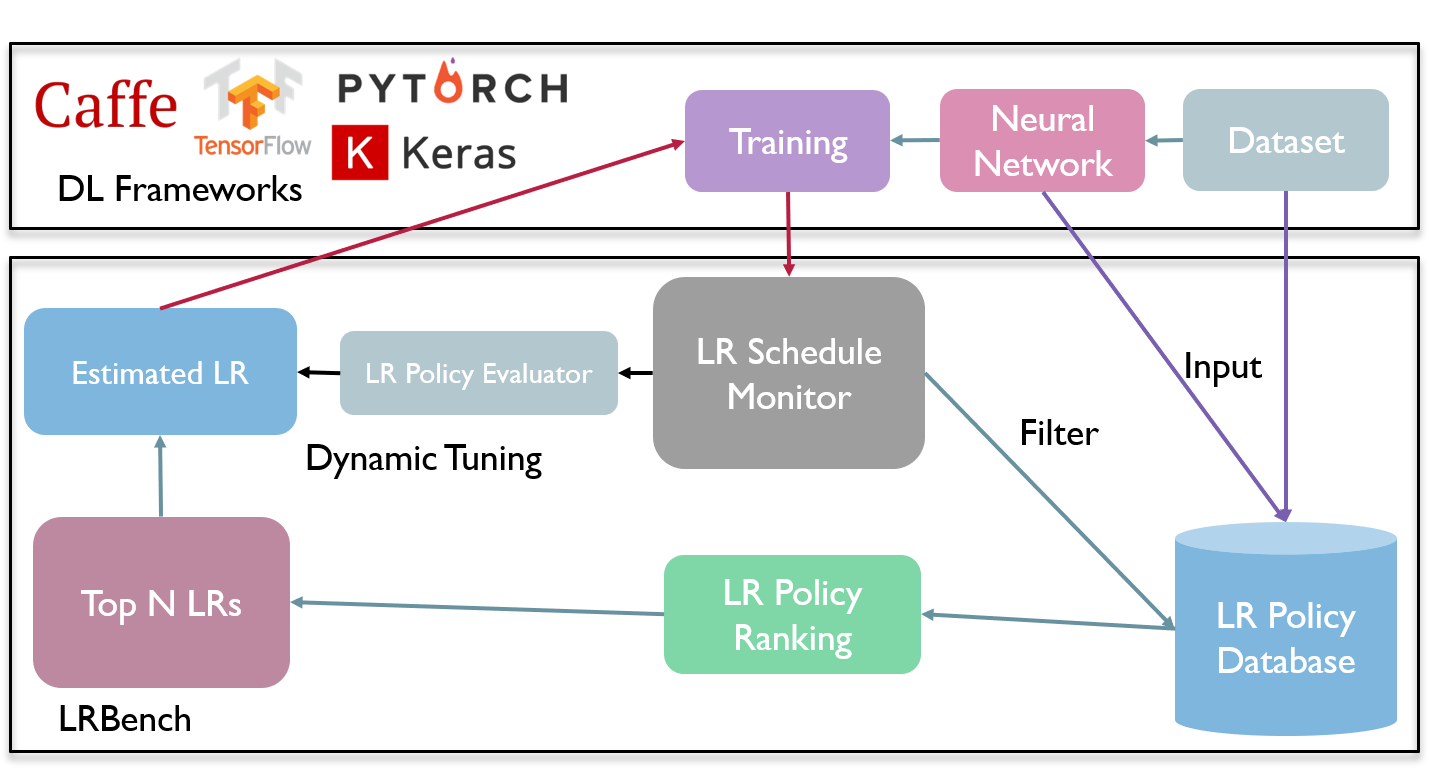}
    \caption{\small{Architecture of LRBench}}
    \label{fig:overall-architecture}
\vspace{-5mm}
\end{figure}

\section{LRBench} \label{section:lrbench}
LRBench is an LR benchmarking system that provides automated or semi-automated tuning and optimization for finding and selecting a good LR policy when DNN developers or end-users have chosen the dataset and the DNN model for training.
LRBench consists of four main functional components as shown in Figure \ref{fig:overall-architecture}: (1) The LR schedule monitor tracks the number of iterations and triggers the LR value update when the update schedule is met. (2) The LR policy evaluator estimates the good LR value range, the step size for LR update schedule, and other utility parameters for evaluating, comparing LR policies and dynamically tuning LR values. (3) The LR policy ranking algorithm will select those top $N$ ranked LR policies based on the empirical measurement results for specified utility, cost and robustness metrics. (4) The LR database, implemented with PostgreSQL, stores the empirical LR policy tuning results organized based on the specific deep learning framework used, the specific neural network model chosen from the DNN library, the learning task type (such as classification of $M$ tasks, $M$ can be 2, 10, 100, 1000), and the specific dataset used for training, such as MNIST, CIFAR-10, CIFAR-100, and ImageNet.

{\bf How to use LRBench?} LRBench is designed with modularity and flexibility for ease of use. We implemented each component of LRBench as an independent module. For example, the LR module implements all LR functions while the LR policy database module handles all database related operations. On the one hand, users may use a single module, such as the LR module to obtain the LR values and apply those values to different DL frameworks. It will avoid the cost for modifying the DL framework dependent modules, such as the monitor module. On the other hand, users can also orchestrate all the components of LRBench and specify the specific tuning process to benefit from the full functionality of efficient LR tuning. This design facilitates the wide applicability and easy extendibility of LRBench. In addition, the LR policy database will be provided as a public platform for end-users to collect and share a rich set of experimental results and benefit from it. Besides, in order to enable reproducibility of our experimental results, the datasets and the source codes used for this study are made available on GitHub  (\url{https://github.com/git-disl/LRBench}).

Given a new learning task on a new dataset and a neural network model used for model training, LRBench will first search its LR policy database, if the related records are found, such as the training results on the same or a similar dataset or neural network, LRBench will use the stored LR ranges as a starting point. Otherwise, LRBench will perform an initial LR range test based on popular heuristics, such as scaling by 10, to determine the good LR ranges, and to reduce the search space significantly. Then based on different training goals, i.e., high accuracy or low cost, LRBench will tune the settings of a chosen LR function in finer granularity. 

The first prototype of LRBench is implemented in Python on top of Caffe~\cite{caffe,caffe-code}. We implement all 13 LR functions and 8 evaluation metrics. We have open sourced the LRBench on GitHub, which can be plugged into or extended for other open source deep learning frameworks, e.g., TensorFlow, Caffe, PyTorch, Keras.

\begin{figure*}[h!]
\centering
    \subfloat[Overall ($k_0$=0$\sim$0.1)]{
    \includegraphics[width=0.33\textwidth]{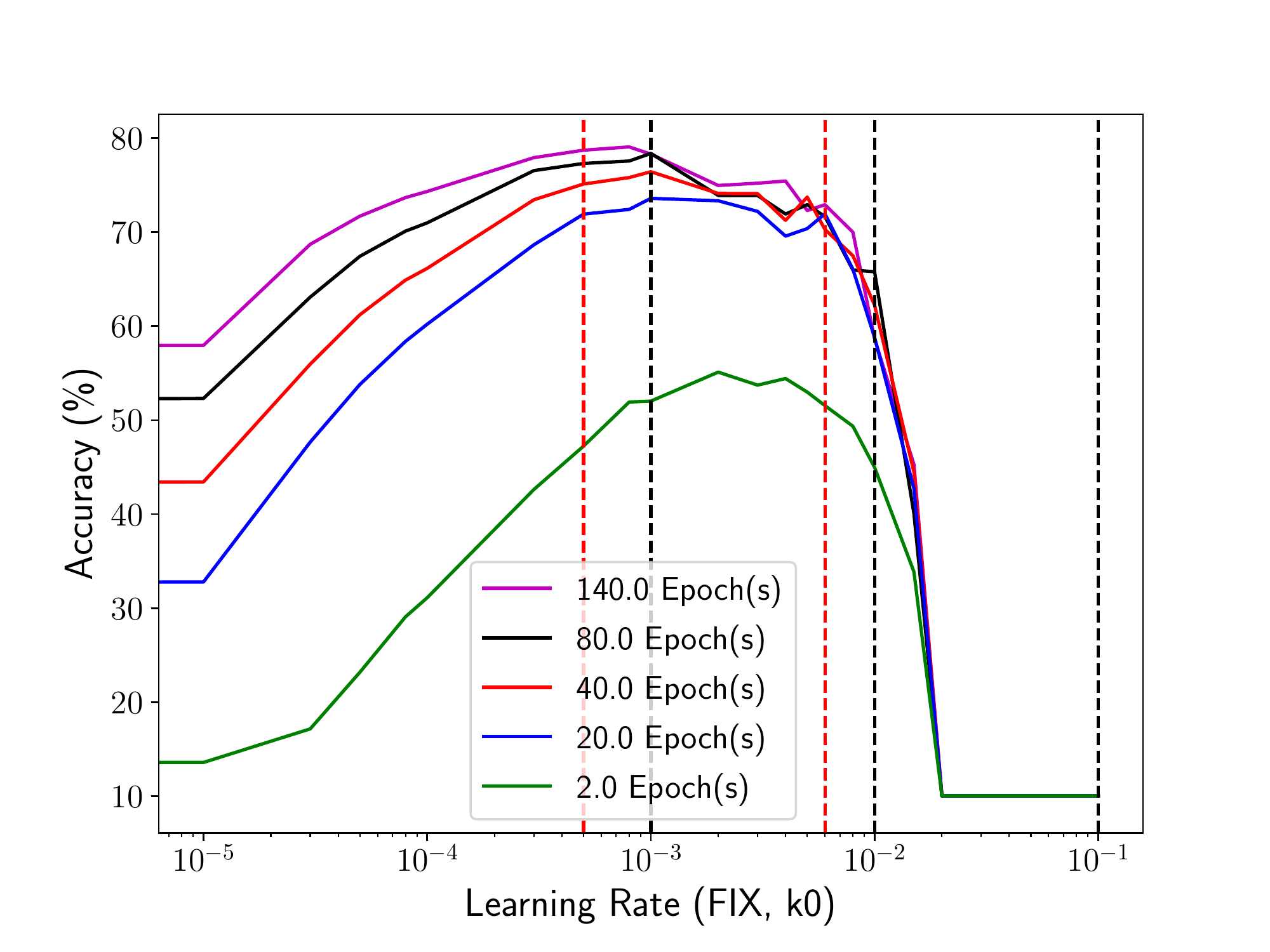}
    \label{fig:acc-k0-cifar10-fix-overall}
    }
    \subfloat[\small{Accuracy Zoom-in ($k_0$=0.00005$\sim$0.01)}]{
    \centering
    \includegraphics[width=0.33\textwidth]{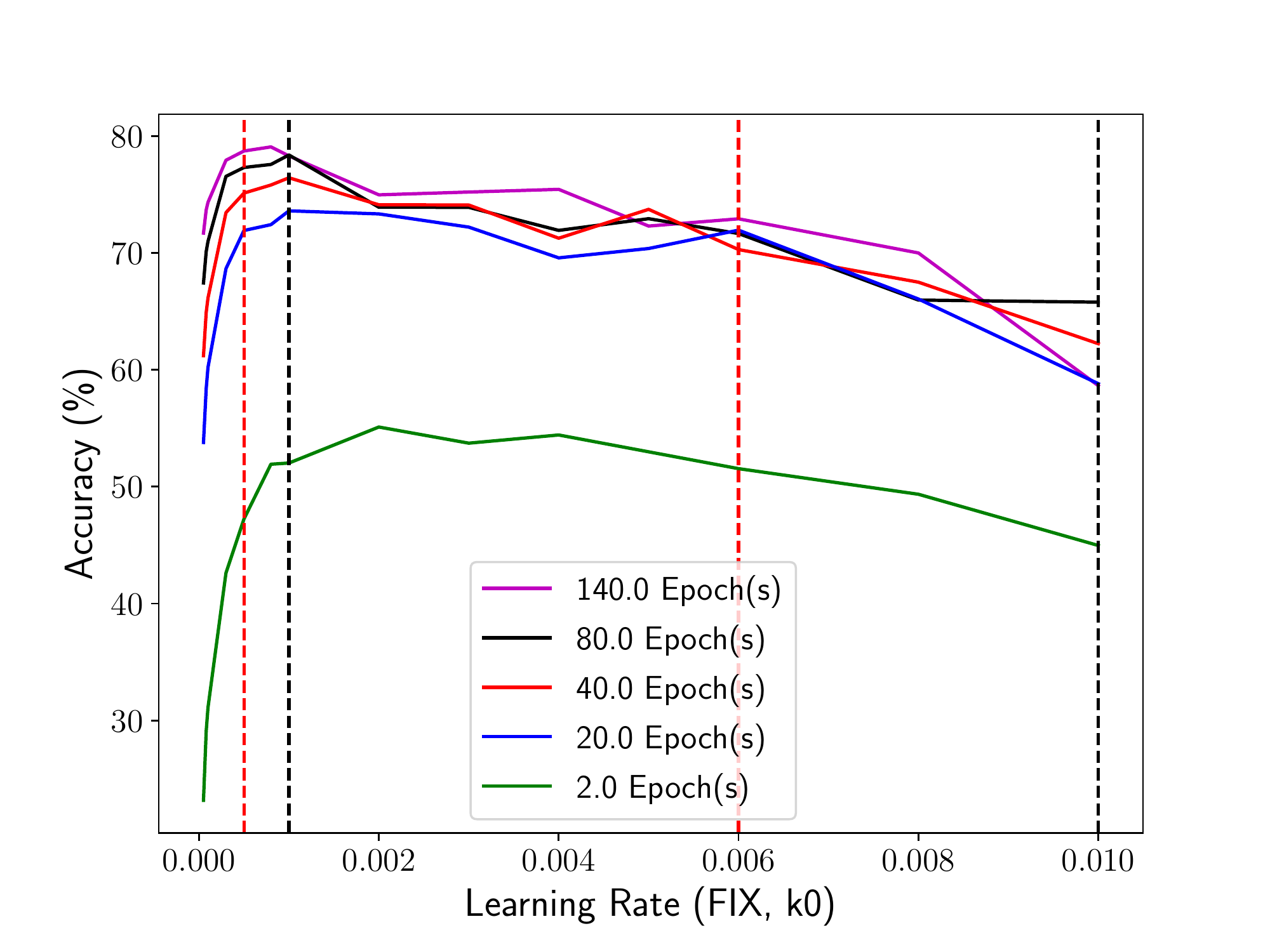}
    \label{fig:acc-k0-cifar10-fix-zoom-in}
    }
    \subfloat[\small{Error Zoom-in ($k_0$=0.00005$\sim$0.01)}]{
    \centering
    \includegraphics[width=0.33\textwidth]{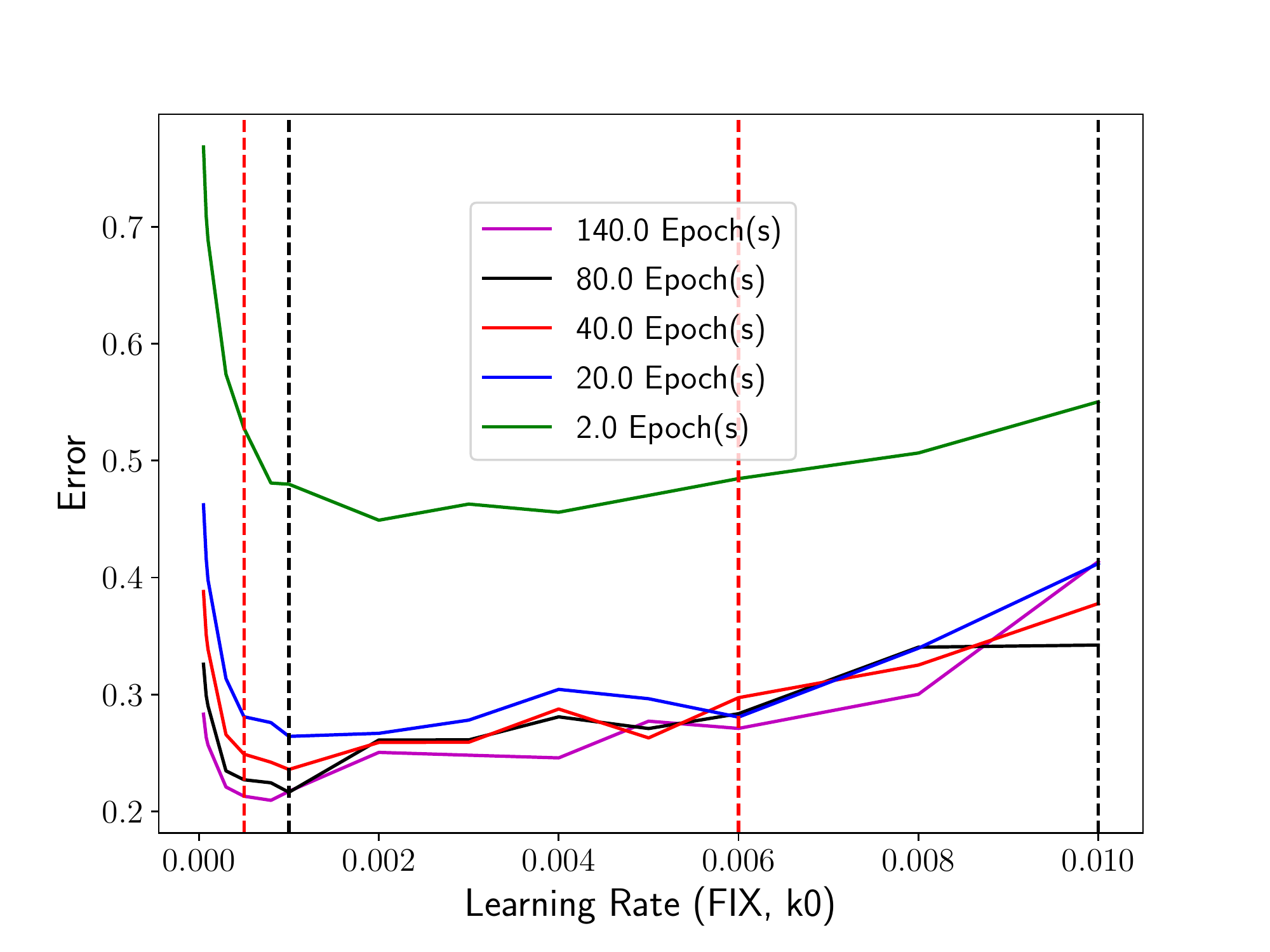}
    \label{fig:error-k0-cifar10-fix-zoom-in}
    }
    \caption{\small{Acc/Error with Varying $k_0$ (FIX, CIFAR-10, CNN3)}}
    \label{fig:k0-cifar10-fix}
    \vspace{-6mm}
\end{figure*}

Most LR functions involve two important sets of parameters: (1) the initial LR value or the minimum and maximum LR boundary values, and (2) the LR value update schedule. Given the large number of possible choices for choosing fixed or variable step sizes, we incorporate some practical heuristics. For example, most of DNN training uses mini-batches. The number of iterations for an epoch is calculated by dividing the total number of training samples by the {\em batch size}. CIFAR-10 has 50,000 training images. With the mini-batch size of 100, an epoch is 500 iterations (50,000/100). An easy way to set the LR update schedule is to use a multiple of the \#Iterations per epoch, because training in one epoch allows the DNN to learn over the entire dataset once and a good DNN training usually requires to repeatedly train the model on the entire dataset multiple times.


\section{Experimental Results and Analysis}
We conduct experiments using LRBench to show how the LR tuning is performed and the impact of datasets and DNN models on the utility, cost and robustness metrics of various LR policies. For each pair of the dataset and neural network, i.e., CIFAR-10 with CNN3, we only vary the default setting of LR value and keep the default settings of all other hyper-parameters (see Table \ref{table:default-training-parameters}). MNIST consists of 70,000 gray-scale images of ten handwritten digits, each image is $28\times28\times1$ in size with 60,000 images for training and the remaining 10,000 images for testing. CIFAR-10 consists of 60,000 colorful images of 10 classes, each is $32\times32\times3$ in size. Similarly, for CIFAR-10, 50,000 images are used for training with the remaining 10,000 images for testing. Different neural networks prefer different default training parameters, even for the same dataset, CIFAR-10. Specifically, LeNet and CNN3 choose 100 as the batch size and SGD as the optimizer while ResNet-32 uses 128 and Nesterov instead. These three neural networks differ in their hyperparameters, their weight initialization algorithms and weight decay parameters, though they all set the momentum as 0.9 for gradient descent.
\begin{table}[h!]
\vspace{-4mm}
\centering
\caption{\small{Default Training Parameters}}
\label{table:default-training-parameters}
\scalebox{1.0}{
\small
\begin{tabular}{|c|c|c|c|}
    \hline
    Framework & \multicolumn{3}{c|}{Caffe+LRBench} \\
    \hline
    Dataset & MNIST & \multicolumn{2}{c|}{CIFAR-10} \\
    \hline
    Neural Network & LeNet & CNN3 & ResNet-32 \\
    \hline
    \#Training Samples & 60,000 & \multicolumn{2}{c|}{50,000} \\
    \hline
    \#Testing Samples & 10,000 & \multicolumn{2}{c|}{10,000} \\
    \hline
    \#Max Iterations & 10,000 & 70,000 & 64,000 \\
    \hline
    Batch Size & 100 & 100 & 128 \\
    \hline
    Optimizer & SGD & SGD & Nesterov \\
    \hline
    Weight Initialization & xavier\cite{xavier} & gaussian & MSRA\cite{resnet} \\
    \hline
    Weight Decay & 0.0005 & 0.004 & 0.0001\\
    \hline
    Momentum & 0.9 & 0.9 & 0.9 \\
    \hline
\end{tabular}
}
\vspace{-2mm}
\end{table}

All experiments are conducted on an Intel Xeon E5-1620 server with one Nvidia GeForce GTX 1080 Ti (11GB) GPU, installed with Ubuntu 16.04 LTS, CUDA 8.0 and cuDNN 6.0.

\subsection{LR Refinement Value Range ($k_0$, $k_1$)}
$k_0$, $k_1$ represent the learning rate range and an LR policy starts from $k_0$. If $k_0=k_1$, it is the fixed learning rate (FIX). For triangle and sin based CLRs ($k_0 < k_1$), each cycle starts with the minimum value ($k_0$), and first increases at each iteration to follow the respective $g(t)$ (as shown in Table~\ref{table:lr}) until it reaches the maximum bound ($k_1$), then the LR value starts to decrease until it drops to the minimum bound ($k_0$). The next cyclic schedule starts to repeat until the training stop condition is reached, e.g., the pre-defined total number of iterations is completed or a pre-defined accuracy threshold is met. For the cosine CLR ($k_0 > k_1$), it starts with the maximum value ($k_0$) and decays the initial value to the minimum value ($k_1$) in the first half cycle and then increases the LR value to the maximum bound to complete one cycle.

\begin{table*}[h!]
\caption{\small{LRBench Recommendations of Good LR Policies for CIFAR-10 (ResNet-32)}}
\label{table:lr-recoomendation-cifar10-resent}
\addtocounter{table}{-1}
\vspace{-2mm}
\centering
\subfloat[FIX]{
\scalebox{1.0}{
\small
\begin{tabular}{|c|c|c|c|}
\hline
LR                   & $k_0$    & \#Iterations & Highest Acc (\%) \\ \hline
\multirow{3}{*}{FIX} & 0.1   & 61000        & \textbf{86.08}      \\ \cline{2-4} 
                     & 0.01  & 50000        & 85.41      \\ \cline{2-4} 
                     & 0.001 & 48000        & 82.66      \\ \hline
\end{tabular}
} 
\label{table:acc-lr-cifar10-fix-resnet}
} 
\subfloat[STEP]{
\scalebox{1.0}{
\small
\begin{tabular}{|c|c|c|c|c|c|}
\hline
LR                     & $k_0$  & $\gamma$ & $l$     & \#Iterations & Highest Acc (\%) \\ \hline
\multirow{3}{*}{STEP} & 0.1 & 0.85  & 1000  & 40000        & 89.76      \\ \cline{2-6} 
                       & 0.1 & 0.85  & 5000  & 61000        & \textbf{91.10}       \\ \cline{2-6} 
                       & 0.1 & 0.85  & 10000 & 57000        & 87.43      \\ \hline
\end{tabular}
} 
\label{table:acc-l-cifar10-step-resnet}
} 
\vspace{-5mm}
\addtocounter{table}{1}
\end{table*}

\begin{table*}[h!]
\caption{\small{Acc with Varying LR parameters (CIFAR-10, CNN3)}}
\label{table:acc-lr-cifar10}
\addtocounter{table}{-1}
\vspace{-2mm}
\centering
\subfloat[Varying $k_0$ and $k_1$]{
\scalebox{1.0}{
\small
\begin{tabular}{|c|c|c|c|c|}
\hline
LR & $k_0$      & $k_1$    & \#Iterations & Highest Acc (\%) \\ \hline
\multirow{10}{*}{TRI2} & 0.00005 & 0.006 & \textbf{67000}        & \textbf{80.86}            \\ \cline{2-5}
& 0.0001  & 0.01  & 70000        & 80.78            \\ \cline{2-5}
& 0.0005  & 0.006 & 70000        & 80.70             \\ \cline{2-5}
& 0.00001 & 0.006 & 56500        & 80.59            \\ \cline{2-5}
& 0.00006 & 0.006 & 69750        & 80.51            \\ \cline{2-5}
& 0.00008 & 0.006 & 65250        & 80.50             \\ \cline{2-5}
& 0.00025 & 0.006 & 69500        & 80.41            \\ \cline{2-5}
& 0.0002  & 0.006 & 68500        & 80.34            \\ \cline{2-5}
& 0.00002 & 0.006 & 68750        & 80.33            \\ \cline{2-5}
& 0.0001  & 0.006 & 53250        & 80.29            \\ \hline
\end{tabular}
} 
\label{table:acc-k0-k1-tri2-cifar10}
} 
\subfloat[Varying $l$]{
\scalebox{1.0}{
\small
\begin{tabular}{|c|c|c|c|c|c|}
\hline
LR                     & $k_0$  & $\gamma$ & $l$     & \#Iterations & Highest Acc (\%) \\ \hline
\multirow{10}{*}{STEP} & 0.001 & 0.85  & 1000  & 37750        & 76.51      \\ \cline{2-6} 
                      & 0.001 & 0.85  & 2000  & 64750        & 79.20      \\ \cline{2-6} 
                      & 0.001 & 0.85  & 3000  & 69250        & 79.89      \\ \cline{2-6} 
                      & 0.001 & 0.85  & 4000  & 68500        & 79.91      \\ \cline{2-6} 
                      & 0.001 & 0.85  & 5000  & 65250        & 79.68      \\ \cline{2-6} 
                      & 0.001 & 0.85  & 6000  & 68500        & 79.80      \\ \cline{2-6} 
                      & 0.001 & 0.85  & 7000  & \textbf{66750}        & \textbf{80.08}      \\ \cline{2-6} 
                      & 0.001 & 0.85  & 8000  & 63250        & 79.85      \\ \cline{2-6} 
                      & 0.001 & 0.85  & 9000  & 69750        & 80.06      \\ \cline{2-6} 
                      & 0.001 & 0.85  & 10000 & 69250        & 79.77      \\ \hline
\end{tabular}
} 
\label{table:acc-l-cifar10-step}
} 
\vspace{-5mm}
\addtocounter{table}{1}
\end{table*}

{\bf FIX.\/}
For the fixed learning rate, a common practice is to start with a value that is not too small, e.g., 0.1, and then exponentially lower it to get the smaller constant values, such as 0.01, 0.001, 0.0001. 
Figure \ref{fig:k0-cifar10-fix} shows the results of the set of experiments by varying the fixed LR values in a base-10 log scale (x-axis) for CNN3 training on CIFAR-10. The three black vertical dashed lines represent $k_0$ = 0.001, 0.01 and 0.1, while the two red vertical dashed lines represent $k_0$=0.0005 and 0.006, and different colors of curve lines correspond to the accuracy measurements at different \#epochs. 
Figure \ref{fig:acc-k0-cifar10-fix-overall},\ref{fig:acc-k0-cifar10-fix-zoom-in} measure the test accuracy (y-axis) and Figure \ref{fig:error-k0-cifar10-fix-zoom-in} measures the test error (y-axis).
We observe that the proper LR ranges for CIFAR-10 is $[0.0005, 0.006]$. From Figure \ref{fig:acc-k0-cifar10-fix-overall}, the accuracy decreases much faster after $k_0=0.006$, thus it is good to set the maximum LR around 0.006. Similarly, by zooming into $[0.00005, 0.01]$ with Figure~\ref{fig:acc-k0-cifar10-fix-zoom-in} and~\ref{fig:error-k0-cifar10-fix-zoom-in}, we can see more clearly that small LRs also achieved decent accuracy, such as $k_0=0.0005$ (the leftmost red vertical dashed lines), showing $[0.0005, 0.006]$ is a good LR range for CNN3 on CIFAR-10. This also provides a good explanation of the default constant LR value of 0.001 in Caffe (CNN3) for CIFAR-10. To further validate our LR range selection method, we use the fixed LR=0.01 for CNN3 training on CIFAR-10, since 0.01 is larger than 0.001 and it is the default setting for LeNet on MNIST in Caffe. However, it achieved much lower accuracy (69.63\%) than the accuracy of 78.62\% with the constant LR of 0.001.

{\bf Benefits of LRBench.\/} From Figure \ref{fig:k0-cifar10-fix}, we also observe and learn the benefits of using LRBench: (1) With five different \#Epochs (five accuracy curves), the general trend of accuracy over different LR values shows similar patterns. This indicates that instead of manually performing trials and errors, LRBench can automate the LR policy selection process by only comparing a range of LR values for a few different \#Epochs instead of enumerating large number of possibilities over the default total \#iterations (Caffe default for CIFAR-10 is 140 epochs or 70,000 iterations). (2) Similarly, LRBench can avoid bad LR policies through an automated benchmarking process, significantly reducing the management cost of manual trial-and-error operations. For example, by investigating the relationship between accuracy and the proper LR range, choosing the constant LR value from a good LR range will reduce the search space by 99.41\% ($(1-(0.006-0.00005))\times100\%/1$) for CIFAR-10 w.r.t. using the domain of $[0, 1]$ as the search space for finding a good constant LR. 
(3) LRBench uses several heuristics learned from empirical observations. For example, although the domain of LR values is $[0, 1]$, most of existing deep learning frameworks uses constant LR values from 0.1 to 0.0001. Thus, we could start the LR from a relatively large value, such as 0.1, and then reduce it by 10 each time until the persistent dropping of accuracy is observed. This heuristic can be a good guideline when a new DNN model is used to train on CIFAR-10, such as ResNet-32.
Table~\ref{table:acc-lr-cifar10-fix-resnet} shows the highest accuracy and the corresponding \#Iterations for training the ResNet-32 on CIFAR-10 for the default 64,000 iterations. Since $k_0=0.1$ achieved the highest accuracy (86.08\%), it is used as the initial LR value for CIFAR-10 on ResNet-32. We defer the discussion on Table~\ref{table:acc-l-cifar10-step-resnet} to later.

{\bf Cyclic LR --- TRI2.\/} 
We next choose TRI2 to study the impact of $k_0$ and $k_1$. For TRI2, we have $k_0<k_1$. Table \ref{table:acc-k0-k1-tri2-cifar10} shows the results for CNN3 on CIFAR-10 with the highest accuracy and the associated \#Iterations (cost) in bold. The top 10 ($N=10$) results ordered by the highest accuracy are included. LRBench will first determine the search space for good value ranges of TRI2 by leveraging the good value range learned from FIX $[0.0005, 0.006]$. In general, a $k_0$ smaller than 0.0005 and a relatively larger $k_1$ will not miss high accuracy scenarios. For CNN3 on CIFAR-10, we achieve higher accuracy with $k_0 \in (0.00001, 0.0001)$ and $k_1\approx0.006$ compared to the default of Caffe ($k_0=0.001$). This is because $k_0$ and $k_1$ play different roles and exhibit different characteristics during training. For TRI2, $k_1$ as the maximum LR value should be larger than the constant LR value for FIX (i.e., 0.001 for CNN3 on CIFAR-10). 
LRBench utilizes the good value range $[0.0005, 0.006]$ from FIX to determine a reasonable upper bound choice, which could be 0.006 or 0.06 for example. On the other hand, given $k_0$ is the lower bound and minimum LR value and $LR_{TRI2} \ge k_0$ (recall Formula \ref{formula:lr}), a smaller $k_0$ than $0.0005$ will help the model to converge and also not to miss high accuracy cases. Again, we could choose to start $k_0$ with 0.00005, reducing 0.0005 by 10, or 0.000005 reducing it by $10^2$. For CNN3 on CIFAR-10, the LR range of $[0.00005, 0.006]$ gives the highest accuracy of 80.86\%, though the range of $k_0=0.0005, k_1=0.06$ is a good range for TRI2 as well, achieving accuracy of 80.78\%, ranked the third in our experiments. This once again shows the benefit of using LRBench over manual tuning, because it can reduce the good LR policy search space significantly by learning from experience while using automation to alleviate the manual management cost of multiple trials. 
Finally, we would like to note that for a new learning task, say training ResNet-32 on CIFAR-10, Caffe does not provide any default LR policy. Using LRBench, we can first search for a good $k_0, k_1$ pair, by selecting a large $k_1$, such as 0.3, 0.6, 0.9, and a small $k_0$, such as 0.01, 0.001, 0.0001, scaled from the FIX LR value of 0.1 (recall Table~\ref{table:acc-lr-cifar10-fix-resnet}). Table \ref{table:acc-k0-k1-cifar10-tri2-resnet} shows the results. With only a few trials, LRBench can identify a good range of $k_0$=0.0001, $k_1$=0.9 for TRI2, offering high accuracy of 91.87\%.

\begin{table}[h!]
\vspace{-4mm}
    \centering
    \caption{\small{LR Recommendation for CIFAR-10 (ResNet-32)}}
    \label{table:acc-k0-k1-cifar10-tri2-resnet}
    \begin{tabular}{|c|c|c|c|c|}
    \hline
    LR & $k_0$     & $k_1$  & \#Iterations & Highest Acc (\%) \\ \hline
    \multirow{6}{*}{TRI2} & 0.0001 & 0.9 & 44000        & \textbf{91.87}            \\ \cline{2-5} 
    & 0.001  & 0.9 & 60000        & 91.61            \\ \cline{2-5} 
    & 0.001  & 0.6 & 55000        & 91.45            \\ \cline{2-5} 
    & 0.0001 & 0.6 & 36000        & 91.43            \\ \cline{2-5} 
    & 0.001  & 0.3 & 32000        & 90.59            \\ \cline{2-5} 
    & 0.0001 & 0.3 & 31000        & 90.58            \\ \hline
    \end{tabular}
    \vspace{-4mm}
  \end{table}

\subsection{LR Value Update Schedule ($l$)}
For multi-step decaying LR functions, an LR policy needs to include an LR value update schedule ($l$), which specifies the number of times and each time at which specific \#Iterations that the LR value should be updated. 
Based on the heuristic of using a multiple of \#Epochs (recall the heuristic in Section \ref{section:lrbench}), LRBench only needs to perform an empirical study on a small set of schedule choices to identify the top ranked LR policies. Note that given a dataset and a DNN model, if the FIX LR policy is already in the LRBench database, we can leverage it to set the initial LR value for a decaying LR function. Otherwise, we will employ the same process to first find a good FIX LR policy and set the initial LR value $k_0$. Thus, in this section, we focus our discussion on the LR value update schedule with STEP and NSTEP as our case studies.

\begin{figure}[h!]
\vspace{-2mm}
\centering
    \includegraphics[trim=10 5 5 10, clip,width=0.4\textwidth]{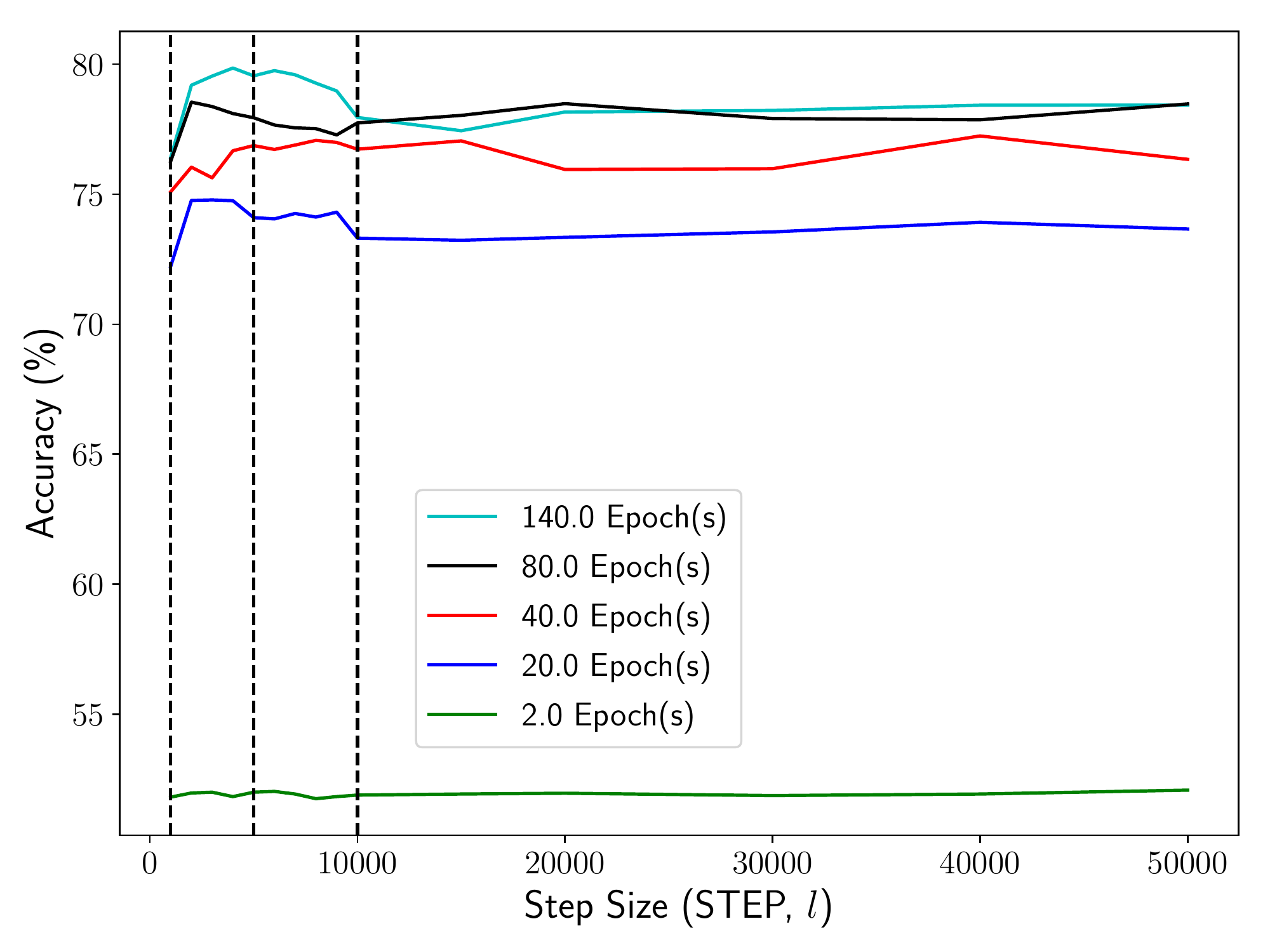}
\caption{\small{STEP (CIFAR-10, CNN3)}}
\label{fig:acc-l-cifar10-step}
\vspace{-2mm}
\end{figure}
{\bf STEP.\/}
STEP is a decaying LR function that reduces the LR value periodically at a fixed step size $l$, defined by \#Iterations. Figure \ref{fig:acc-l-cifar10-step} and Table \ref{table:acc-l-cifar10-step} show the STEP experimental results with varying $l$ for CNN3 training on CIFAR-10 with the same initial $k_0=0.001$ and the same $\gamma=0.85$. The three black vertical dashed lines represent $l=1000,~5000$ and $10,000$ respectively. We observe that (1) different settings of $l$ show small impact on accuracy for CNN3 training on CIFAR-10, compared to the impact of the value range of $k_0, k_1$, except a small peak for $l\in[1000, 10000]$ as Figure \ref{fig:acc-l-cifar10-step} shows. Therefore, we only show the accuracy measurements for $l\in[1000, 10000]$ on Table \ref{table:acc-l-cifar10-step}. The best accuracy is 80.08\% when $l=7000$, which is also chosen by LRBench for training CNN3 on CIFAR-10. (2) Given $l$ determines the \#Steps ($= (\#Iterations / l)-1$), a smaller step size $l$ implies more steps, i.e., more frequent reduction of the LR value, over the total default number of training iterations, and thus it may slow down the model training due to small LR values. Consider training 70,000 iterations for CNN3 on CIFAR-10 with step size $l=1000$, the final LR will be $0.001\times0.1^{70000/1000-1}$, which is almost approaching 0. Hence, we argue that the STEP decaying LR policy should add a lower bound LR value $k_1$ in addition to $l$ to avoid this issue caused by a small step size for a training algorithm that requires large \#Iterations. (3) For training a new DNN model on CIFAR-10, say ResNet-32, LRBench can leverage the good $l$ schedules for CNN3 on CIFAR-10 as the starting $l$. Assume we use $l=1000, 5000, 10000$, Table \ref{table:acc-l-cifar10-step-resnet} shows that $l=5000$ achieved the highest accuracy.

  \begin{table}[h!]
  \vspace{-3mm}
    \centering
    \caption{\small{NSTEP with Different $\gamma$ and $l$ (CIFAR-10, CNN3)}}
    \label{table:acc-l-cifar10-nstep}
    \begin{tabu}{|c|c|c|c|}
    \hline
    $\gamma$ & $l$                          & \#Iteration & Highest Acc (\%) \\ \hline
    0.1   & \makecell{\textit{60000},\\\textit{65000}}               & 70000       & 81.51      \\ \hline 
    0.1   & \makecell{30000, 50000,\\60000, 65000} & 60500       & 81.51      \\ \hline 
    \textbf{0.32}   & \makecell{30000, 50000,\\60000, 65000} & 69750       & \textbf{81.76}      \\ \hline
    \end{tabu}
  \end{table}
  
  \begin{figure}[h!]
  \vspace{-4mm}
    \centering
    \includegraphics[trim=10 5 5 10, clip,width=0.45\textwidth]{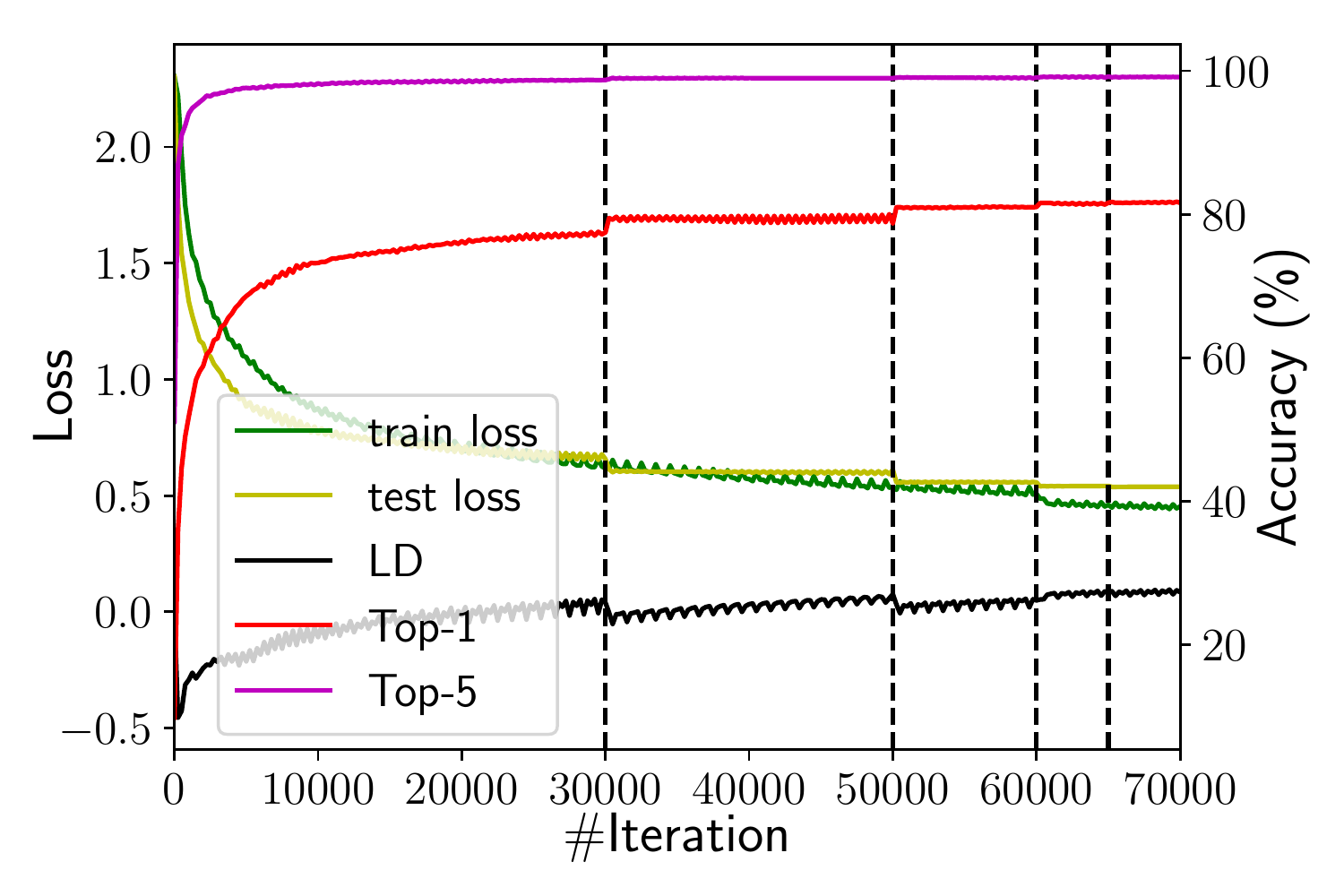}
    \caption{\small{NSTEP (CIFAR-10, CNN3)}}
    \label{fig:acc-loss-cifar10-nstep}
  \end{figure}

\begin{table*}[!h]
\vspace{-4mm}
\caption{\small{Metric Evaluation (MNIST, LeNet)}}
\vspace{-2mm}
\label{table:metric-evaluation-mnist}
\centering
\scalebox{0.9}{
\small
\begin{tabular}{|c|c|c|c|c|c|c|c|c|c|c|c|c|c|}
\hline
LR     & $k_0$   & $k_1$   & $\gamma$   & $p$    & $l$                                                                       & \#Iter & LD     & Top-1 (\%)  & Top-5 (\%) & AC     & CD     & CDAC   & Source  \\ \hline
NSTEP  & 0.01 &      & 0.9     &      & \scriptsize {\begin{tabular}[c]{@{}c@{}}5000,7000,\\8000,9000,9500\end{tabular}} & 10000  & 0.0209 & 99.12$\pm$0.01 & 99.99      & 0.9946 & 0.0362 & 0.0023 & Default \\ \hline
INV    & 0.01 &      & 0.0001  & 0.75 &                                                                         & 8000   & 0.0228 & 99.09$\pm$0.02 & 99.99      & 0.9945 & 0.0369 & 0.0021 & Default \\ \hline
POLY   & 0.01 &      &         & 1.2 &      & 8000  & 0.0210 & 99.11$\pm$0.01 & 99.99  & 0.9937 & 0.0389 & 0.0185 & LRBench \\ \hline
TRI    & 0.01 & 0.06 &         &     & 2000 & 4000  & 0.0166 & 99.28$\pm$0.02 & 100.00 & 0.9948 & 0.0349 & 0.0020 & LRBench \\ \hline
TRI2   & 0.01 & 0.06 &         &     & 2000 & 4000  & 0.0170 & 99.28$\pm$0.01 & 100.00 & 0.9949 & 0.0344 & 0.0018 & LRBench \\ \hline
TRIEXP & 0.01 & 0.06 & 0.99994 &     & 2000 & 4000  & 0.0167 & 99.27$\pm$0.01 & 100.00 & 0.9948 & 0.0353 & 0.0019 & LRBench \\ \hline
SIN    & 0.01 & 0.06 &         &     & 2000 & 4000  & 0.0171 & 99.31$\pm$0.04 & 100.00 & 0.9949 & 0.0350 & 0.0171 & LRBench \\ \hline
SIN2   & 0.01 & 0.06 &         &     & 2000 & 4000  & 0.0166 & \textbf{99.33}$\pm$0.02 & 100.00 & 0.9950 & 0.0342 & 0.0018 & LRBench \\ \hline
SINEXP & 0.01 & 0.06 & 0.99994 &     & 2000 & 4000  & 0.0168 & 99.28$\pm$0.04 & 100.00 & 0.9950 & 0.0340 & 0.0019 & LRBench \\ \hline
COS    & 0.01 & 0.06 &         &     & 2000 & 10000 & 0.0192 & 99.32$\pm$0.04 & 100.00 & 0.9952 & 0.0336 & 0.0192 & LRBench \\ \hline
\end{tabular}} 
\vspace{-2mm}
\end{table*}

\begin{table*}[!h]
\caption{\small{Metric Evaluation (CIFAR-10, CNN3)}}
\label{table:metric-evaluation-cifar10}
\vspace{-2mm}
\centering
\scalebox{0.9}{
\small
\begin{tabular}{|c|c|c|c|c|c|c|c|c|c|c|c|c|c|}
\hline
LR     & $k_0$      & $k_1$    & $\gamma$   & $p$ & $l$                   & \#Iter & LD     & Top-1 (\%)  & Top-5 (\%) & AC     & CD     & CDAC   & Source   \\ \hline
NSTEP  & 0.001   &       & 0.1     &   & \scriptsize{60000, 65000} & 62250  & 0.1758 & 81.61$\pm$0.18 & 99.16      & 0.8695 & 0.1713 & 0.0559 & Default   \\ \hline
TRI    & 0.001   & 0.006 &         &  & 2000 & 68000 & 0.1493 & 79.39$\pm$0.17 & 98.95 & 0.8630 & 0.1848 & 0.0607 & \cite{clr} \\ \hline
TRI2   & 0.001   & 0.006 &         &  & 2000 & 65500 & 0.1881 & 79.75$\pm$0.14 & 98.80 & 0.8652 & 0.1800 & 0.0601 & \cite{clr} \\ \hline
TRIEXP & 0.001   & 0.006 & 0.99994 &  & 2000 & 70000 & 0.1827 & 80.16$\pm$0.12 & 98.85 & 0.8686 & 0.1780 & 0.0545 & \cite{clr} \\ \hline
TRI    & 0.00005 & 0.006 &         &  & 2000 & 68000 & 0.1150 & 81.75$\pm$0.13 & 99.16 & 0.8664 & 0.1812 & 0.0604 & LRBench   \\ \hline
TRI2   & 0.00005 & 0.006 &         &  & 2000 & 70000 & 0.1252 & 80.71$\pm$0.14 & 99.09 & 0.8543 & 0.1779 & 0.0617 & LRBench   \\ \hline
TRIEXP & 0.00005 & 0.006 & 0.99994 &  & 2000 & 68000 & 0.1637 & 81.92$\pm$0.13 & 99.17 & 0.8679 & 0.1723 & 0.0550 & LRBench   \\ \hline
SIN    & 0.00005 & 0.006 &         &  & 2000 & 68000 & 0.1068 & 81.76$\pm$012  & 99.03 & 0.8636 & 0.1840 & 0.0637 & LRBench   \\ \hline
SIN2   & 0.00005 & 0.006 &         &  & 2000 & 70000 & 0.1297 & 80.79$\pm$0.14 & 99.11 & 0.8549 & 0.1781 & 0.0621 & LRBench   \\ \hline
SINEXP & 0.00005 & 0.006 & 0.99994 &  & 2000 & 52000 & 0.1486 & \textbf{82.16}$\pm$0.08 & 99.19 & 0.8632 & 0.1770 & 0.0586 & LRBench   \\ \hline
COS    & 0.00005 & 0.006 &         &  & 2000 & 70000 & 0.1130 & 81.43$\pm$0.14 & 99.02 & 0.8616 & 0.1832 & 0.0605 & LRBench   \\ \hline
\end{tabular}
} 
\vspace{-2mm}
\end{table*}

\begin{table*}[!h]
\caption{\small{Metric Evaluation (CIFAR-10, ResNet-32)}}
\vspace{-2mm}
\label{table:metric-evaluation-cifar10-resnet}
\centering
\scalebox{0.9}{
\small
\begin{tabular}{|c|c|c|c|c|c|c|c|c|c|c|c|c|c|}
\hline
LR     & $k_0$     & $k_1$  & $\gamma$   & $p$ & $l$            & \#Iter & LD     & Top-1 (\%)  & Top-5 (\%) & AC     & CD     & CDAC   & Source         \\ \hline
NSTEP  & 0.1    &     & 0.1     &   & 32000, 48000 & 53000  & 0.3490 & 92.38$\pm$0.04 & 99.77      & 0.9850 & 0.0646 & 0.0073 & \cite{resnet} \\ \hline
TRIEXP & 0.0001 & 0.9 & 0.99994 &   & 2000    & 64000  & 0.3218 & 92.76$\pm$0.14 & 99.74      & 0.9860 & 0.0605 & 0.0059 & LRBench        \\ \hline
SINEXP & 0.0001 & 0.9 & 0.99994 &   & 2000   & 64000  & 0.3116 & \textbf{92.81}$\pm$0.08 & 99.79      & 0.9851 & 0.0639 & 0.0071 & LRBench        \\ \hline
\end{tabular}
} 
\vspace{-6mm}
\end{table*}

{\bf NSTEP.\/}
NSTEP is a multi-step decaying LR function with variable step sizes, aiming to update the LR value when the test accuracy or loss fails to improve. 
Table \ref{table:acc-l-cifar10-nstep} shows the the highest Top-1 accuracy and the corresponding \#Iterations for training CNN3 on CIFAR-10 over the entire default \#Iterations with the default $k_0=0.001$. Given the third configuration achieved the highest accuracy of 81.76\%, Figure \ref{fig:acc-loss-cifar10-nstep} further shows the accuracy and loss measurement with $\gamma=0.32$ and $l=[30000, 50000, 60000, 65000]$). We make two remarks. First, the LR value decay should be triggered at the appropriate time to facilitate model training. The four vertical dashed lines in Figure \ref{fig:acc-loss-cifar10-nstep} show the loss and accuracy measurements when $l=[30000, 50000, 60000, 65000])$. Clearly at these four steps, either accuracy or loss is dropped, which is a good timing for triggering an LR value update using the corresponding decay function. Second, $\gamma$ and $l$ are correlated and should be adjusted accordingly. Given $l$ determines the \#Steps, a large number of steps will make the final LR smaller. For example, with $k_0=0.001$ as the starting point and $\gamma=0.1$, the 4 steps of update will result in LR=0.0000001 ($0.001\times0.1^4$), which is again too small for training (the second configuration in Table \ref{table:acc-l-cifar10-nstep}). If we only use two steps by setting $l=60000, 65000$, it will result in LR=0.00001 ($0.001\times0.1^2$) for the last phase of the training. However, when we enlarge $\gamma$ to 0.32, the 4 steps of update will result in 0.00001 ($0.001\times0.32^4$) which is the same for the 2 step case with $\gamma=0.1$, not too small for the last training phase. This also explains that the 4 steps with $\gamma=0.32$ (the third configuration in Table \ref{table:acc-l-cifar10-nstep}) achieved the highest accuracy.

\subsection{Evaluating LR Policies: Utility, Cost and Robustness}
We study 
the impact of our proposed metrics on selecting good LR policies for training LeNet on MNIST, CNN3 and ResNet-32 on CIFAR-10. 
Tables \ref{table:metric-evaluation-mnist}, \ref{table:metric-evaluation-cifar10} and \ref{table:metric-evaluation-cifar10-resnet} show the results.
We marked the source of the LR parameters as Default when it is the same as those recommended by Caffe, or existing studies~\cite{clr,resnet} or LRBench and we primarily show the top LR policies recommended by LRBench. For Top-1 accuracy, we use the averaged values from 5 repeated experiments with the mean$\pm$stddev. In general, LRBench is able to identify good LR policies efficiently and can improve the accuracy compared to that of the default LR policy with lower training cost.

{\bf MNIST (LeNet).\/} Table \ref{table:metric-evaluation-mnist} shows the experimental results on Caffe with $10,000$ iterations as the default. We make three observations.

{\it First,} LRBench successfully identified SIN2 as the winner for the highest accuracy with the lowest cost. Compared to the default LR policies by Caffe (NSTEP and INV), SIN2 improved the accuracy significantly to 99.33\% by 0.21\% and 0.24\% with much lower cost, that is 4,000 iterations for SIN2, reducing the cost by more than half, comparing to 10,000 iterations for NSETP and 8,000 iterations for INV.

{\it Second,} the 7 CLRs recommended by LRBench produce higher accuracy with Top-1 (99.27\%$\sim$99.33\%), lower CD (confidence deviation), and lower CDAC (confidence deviation across classes). In particular, the loss difference (LD) for CLRs is lower than decaying LRs, demonstrating their good resistance to over-fitting. Moreover, CLRs achieve the higher accuracy with fewer iterations, thus lower cost. To achieve the accuracy of over 99.27\%, TRI, TRI2, TRIEXP, SIN, SIN2 and SINEXP only takes 4000 iterations, while the Caffe default LRs (NSTEP, INV) fail to achieve such high accuracy even after over 8000 iterations.

{\it Third,} the proposed metrics provide a comprehensive comparison of various LR policies in addition to the accuracy measurement. For example, these experiments show that TRI, TRI2 and SINEXP achieved the same accuracy (99.28\%), but they differ in LD, AC, CD and CDAC. For example, SINEXP achieved the higher AC and lower CD, indicating better classification confidence and stability. 

In summary, SIN2 is identified by LRBench as the best CLR policy with the highest accuracy for LeNet on MNIST at the reduced cost of only half of the default training time in Caffe.

{\bf CIFAR-10 (CNN3).\/}
Table \ref{table:metric-evaluation-cifar10} shows the experimental results with Caffe and its default total training of $70,000$ iterations. We make three interesting observations.

{\it First,} in addition to the NSTEP with accuracy 81.61\% as the Caffe default LR, LRBench successfully identified another 4 LR policies: SINEXP (LRBench, 82.16\%), TRIEXP (LRBench, 81.92\%), SIN (LRBench, 81.76\%) and TRI (LRBench, 81.75\%). SINEXP (LRBench) significantly improves the accuracy by 0.55\% with a training cost reduction of 10,250 iterations (14.6\% of the default 70,0000 iterations), compared to the Caffe default NSTEP. In addition, the ACs of the top 5 accuracy LR policies (NSTEP and the four LRs identified by LRBench) are in the close range of 0.8632$\sim$0.8695, higher than other LRs, indicating better classification confidence.

{\it Second,} using other metrics in LRBench, we also show that the CDs (confidence deviations) of SINEXP, TRIEXP, SIN and TRI (0.1723$\sim$0.1840) are higher than NSTEP (0.1713). This shows that even though SINEXP, TRIEXP, SIN and TRI achieved higher accuracy of 81.75\%$\sim$82.16\%, their classification confidence stability is slightly lower than NSTEP (81.61\% accuracy).

{\it Third,} we show that LRBench can be useful for improving the default or recommended LR settings in the literature. For instance, the settings of CLR chosen by LRBench show remarkable advantages over the default triangle based CLRs recommended by Smith in his original paper~\cite{clr}. TRI, TRI2 and TRIEXP chosen by LRBench all outperform the corresponding CLRs recommended in~\cite{clr} w.r.t. Top-1 and Top-5 accuracy and loss difference (LD), 
showing higher utility and better resistance to over-fitting.
In short, CLRs chosen by LRbench show significant performance improvement over the Caffe default NSTEP and the default triangle based CLRs in~\cite{clr}.

{\bf CIFAR-10 (ResNet-32).\/}
ResNet is known to achieve over 90.00\% accuracy on CIFAR-10, a significant improvement over Caffe's CNN3. Table \ref{table:metric-evaluation-cifar10-resnet} shows the results of the two best LR policies identified by LRBench. We observe that TRIEXP and SINEXP offer higher Top-1 and Top-5 accuracy and lower LD, CD and CDAC compared to the default chosen by the ResNet original paper in~\cite{resnet}. For Top-1 accuracy, SINEXP achieved 92.81\% compared to the accuracy of 92.38\% by the default LR in ResNet-32 according to~\cite{resnet}.  

We have reported a set of experiments to demonstrate that a good LR policy often depends on dataset and DNN model specific characteristics. LRBench is practical, helpful and can effectively identify good LR policies, prune out bad policies, reduce the manual management cost of choosing good LR policies, evaluating the default LR policy, avoiding bad LR policies, and consequently, increasing the efficiency and the overall performance of a DNN framework like Caffe.

\section{Related Work}
Although the learning rate (LR) is widely recognized as an important hyper-parameter for training neural networks to achieve high test accuracy~\cite{clr,superconvergence,largeminibatch,understanding-lr-blog}, there are few studies to date dedicated to this topic. Recently, \cite{clr} proposed the triangle cyclic learning rates of three forms: TRI, TRI2, and TRIEXP. A subsequent work \cite{superconvergence} further shows that large learning rate values for CLRs can accelerate neural network training, referred to as super-convergence. Another research effort~\cite{largeminibatch} studies the effect of tuning the batch size hyper-parameter together with the learning rate, and suggests adjusting learning rates as a function of the batch size. To the best of our knowledge, our work is the first to present a comprehensive characterization of over a dozen (13) learning rate functions and the role of setting various parameters in a LR function on training and testing accuracy of the DNN models, such as the range parameters, the step parameters, and the value update parameters. To provide fair comparison of different LR functions and different settings of one LR function (we call them LR policies), we also proposed a set of LR measurement metrics in terms of utility, cost and robustness. Furthermore, we design and implement a LR benchmarking tool, LRBench, to help practitioners to systematically evaluate different LR policies and efficiently identify a good LR policy that best fits their DNN training objectives.

Another related research theme is the hyper-parameter search for finding the optimal setting, which is an active research field in the last decade. Existing hyperparameter search tools, such as Hyperopt~\cite{hyperopt}, SMAC~\cite{smac}, and Optuna~\cite{optuna}, use classical and general hyper-parameter search algorithms, e.g., grid search, random search and Bayesian optimization, which can be costly due to the exhaustive search. LRBench is proposed to reduce the cost of hyler-parameter search by limiting or avoiding exhaustive LR search. For instance, by storing empirical LR tuning results, LRBench will recommend good LR policies for a learning task based on the stored metadata and knowledge on the past tuning experience. This may include the similarity metrics to a known dataset, a DNN model, and/or a learning task, combined with the ranking score of existing LR policies for a given DNN and training dataset. For instance, the good LR policies for training CIRFAR-10 could be recommended to a similar color image dataset of 10 labeled classes using a similar DNN model. 
This enables LRBench to serve as a general reference platform for tuning LR functions to find good LR policies and for sharing these training experience and good LR policies from past users to facilitate LR tuning on new datasets and/or new DNN models.

\section{Conclusion}
We have presented a systematic approach for tuning LRs with three unique contributions. First, we provide a comprehensive characterization of 13 existing LR functions and their related LR policies by finding good LR value parameters, LR range parameters and LR epoch-based update schedules. Second, we proposed a set of metrics for evaluating and comparing different LR policies in terms of utility, cost and robustness. We show that these metrics can help users and DNN developers to evaluate the different LR policies and recommendations and find the LR policy that can best fit their DNN training need. Third but not the least, we design and develop LRBench, a LR benchmarking system, and demonstrate via comprehensive experiments to show that LRBench is easy to use and can find good LR policies by determining the good LR value range, and finding the good LR update schedules using our proposed metrics. 
Our ongoing work continues along two directions. On the one hand, we are working on the deployment of LRBench open sourced on GitHub to a set of popular open source DNN frameworks, such as PyTorch, TensorFlow and Keras. On the other hand, we are investigating the complex interactions between a good LR policy and different GD optimizers.

\section*{Acknowledgment}
This work was a part of XDefense umbrella project for robust diversity ensemble against deception. This research is partially sponsored by NSF CISE SaTC (1564097), SAVI/RCN (1402266, 1550379), CNS (1421561), CRISP (1541074), and an REU supplement (1545173). Any opinions, findings, and conclusions or recommendations expressed in this material are those of the author(s) and do not necessarily reflect the views of the National Science Foundation or other funding agencies and companies mentioned above.
\bibliographystyle{IEEEtran}
\bibliography{reference}

\end{document}